\documentclass[10pt,twocolumn,letterpaper]{article}

\usepackage[final]{cvpr}      
\usepackage{amsmath}
\usepackage{amssymb}
\usepackage{booktabs}
\usepackage{url}

\usepackage{times,graphicx,amsmath,amssymb,booktabs,tabulary,multirow,overpic,xcolor}
\usepackage{colortbl}
\usepackage{makecell}
\usepackage{float}

\definecolor{sh_gray}{rgb}{0.84,0.84,0.84}
\definecolor{sh_gray2}{rgb}{1,0.89,0.75}
\definecolor{color3}{rgb}{0.95,0.95,0.95}
\definecolor{color4}{rgb}{0.96,0.96,0.86}
\definecolor{color5}{rgb}{0.90,0.90,0.90}
\usepackage[pagebackref,breaklinks,colorlinks]{hyperref}
\usepackage[capitalize]{cleveref}
\usepackage{multirow}
\crefname{section}{Sec.}{Secs.}
\Crefname{section}{Section}{Sections}
\Crefname{table}{Table}{Tables}
\crefname{table}{Tab.}{Tabs.}
\usepackage{algorithm}
\usepackage{algorithmic}

\begin{document}

\title{OphGLM: Training an Ophthalmology Large Language-and-Vision Assistant based on Instructions and Dialogue}

\author{%
	Weihao Gao$^{1,*}$, Zhuo Deng$^{1}$\thanks{$*$ = equal contribution, $\dagger$ = corresponding author}~, Zhiyuan Niu$^{1}$, Fuju Rong$^{1}$, Chucheng Chen$^{1}$, Zheng Gong$^{1}$, Wenze Zhang$^{1}$, \\ Daimin Xiao$^{1}$, Fang Li$^{1}$, Zhenjie Cao$^{1}$, Zhaoyi Ma$^{2,\dagger}$, Wenbin Wei$^{3,\dagger}$,Lan Ma$^{1,\dagger}$\\
    $^{1}$Shenzhen International Graduate School, Tsinghua University\\
    $^{2}$National Health Commission Capacity Building and Continuing Education Center\\
    $^{3}$Beijing Tongren Eye Center,Beijing Tongren Hospital
    }
\maketitle

%%%%%%%%% ABSTRACT
\begin{abstract}
	Large multimodal language models (LMMs) have achieved significant success in general domains. However, due to the significant differences between medical images and text and general web content, the performance of LMMs in medical scenarios is limited. In ophthalmology, clinical diagnosis relies on multiple modalities of medical images, but unfortunately, multimodal ophthalmic large language models have not been explored to date. In this paper, we study and construct an ophthalmic large multimodal model. Firstly, we use fundus images as an entry point to build a disease assessment and diagnosis pipeline to achieve common ophthalmic disease diagnosis and lesion segmentation. Then, we establish a new ophthalmic multimodal instruction-following and dialogue fine-tuning dataset based on disease-related knowledge data and publicly available real-world medical dialogue. We introduce visual ability into the large language model to complete the ophthalmic large language and vision assistant (OphGLM). Our experimental results demonstrate that the OphGLM model performs exceptionally well, and it has the potential to revolutionize clinical applications in ophthalmology. The dataset, code, and models will be made publicly available at \url{https://github.com/ML-AILab/OphGLM}.
\end{abstract}

%%%%%%%%% BODY TEXT
%\vspace{-8.5mm}

\section{Introduction}
\label{sec:introduction}
The web sources data like web images and their associated captions are abundantly available in recent years. This kind of parallel image-text data is of great potential to be applied in self-supervised vision-language modeling. As demonstrated by multimodal GPT-4 \cite{openai2023gpt4} and open-sourced efforts such as LLaVA \cite{liu2023visual}, generative pretraining has effectively leveraged their performance. In attempting to improve the model’s knowledge of general facts based on multimodal input, several large-scale multimodal models (LMMs) successfully show strong zero-shot task completion performance. The application of LLMs in various user-oriented visual-language tasks (such as image understanding and reasoning) has shown their great potential of developing universal multimodal conversation assistants.

Although LLMs have succeeded in the general domain, their performance is limited in biomedical scenarios. Since biomedical image-text pairs are quite different from general web content, the LLMs trained with web source data may have problems dealing with professional conversations. Therefore, a visual assistant in the general domain may behave like a non-specialist and be able to answer biomedical questions precisely or produce totally incorrect answers or false facts. Though much progress has been made in biomedical visual question answering (VQA), previous methods often formalize questions into classification (e.g., their answers are limited to ones observed in the training set) and are not suitable for open-ended guidance following. Meanwhile, current research is often based on single-modal text data rather than multimodal image-text data. Therefore, although conversational AI has demonstrated great potential in biomedical applications\cite{NEJMsr2214184,nori2023capabilities,ALGTP4BOOK}, more effort is required in this field.

Ophthalmology is a discipline that relies on multimodal diagnosis with various ophthalmology images, including fundus images, OCT, ultra-widefield fundus images, and fundus fluorescence angiography (FFA) images. However, currently, it lacks an available multimodal medical dialogue system constructed based on large-scale language models to assist the diagnosis progress. In this study, we attempt to construct a ophthalmology large language-and-vision assistant (OphGLM) with fundus images as the entry point. With this LLM, we can further explore the possibility of developing universal multimodal conversation assistants of good performance for biomedical and other scenarios.

Our contributions can be summarized as follows:
\begin{itemize}
    \item We use knowledge graphs and real-world medical dialogues to build an instructions and dialogue fine-tuning dataset for ophthalmic diseases using ChatGPT. This improves the authenticity and usability of LLM in medical question-answering in the healthcare field.

    \item Based on fundus images from public datasets, we construct a computer vision model for common disease diagnosis, which classifies diabetic retinopathy, age-related macular degeneration, pathological myopia, and glaucoma. In addition, we also develop a diabetic retinopathy grading model, a diabetic retinopathy lesion segmentation model, and an ophthalmic rare disease fundus diagnosis model, which serve as disease diagnosis models.

    \item We propose a novel ophthalmology large language-and-vision assistant (OphGLM). To the best of our knowledge, it is the first attempt to combine visual models with large language models in ophthamology. The experimental results demonstrate our OphGLM has the great potential in ophthalmology clinical applications.

\end{itemize}

\section{Related Work}
\label{sec:related}
\subsection{Biomedical Chatbots}
Large Language Models (LLMs) have achieved immense success in the general domain, with ChatGPT \cite{openai2023gpt4} being one of the most representative instances. Moreover, certain advancements have been made in research concerning the open-source fine-tuning of these large language models. Inspired by these successful cases in LLMs, a series of biomedical large language models have been successively introduced, including ChatDoctor \cite{yunxiang2023chatdoctor}, Med-Alpaca \cite{han2023medalpaca}, PMC-LLaMA \cite{wu2023pmc}, DoctorGLM \cite{xiong2023doctorglm}, and Huatuo \cite{wang2023huatuo}.Typically, these models start with pre-trained open-source LLMs and are then fine-tuned on domain-specific biomedical instruction-following datasets. The fine-tuned LLMs exhibit considerable application potential within specific biomedical fields, as they are able to perform a variety of tasks including patient dialogue, provision of diagnostic and treatment recommendations, and explanation of medical knowledge.

HuatuoGPT is a biomedical chatbot, incorporating distilled data generated by ChatGPT along with real-world physician response data. Its consistency and reliability have been validated both in automatic and human evaluations. However, HuatuoGPT only accepts text inputs, significantly limiting its application scenarios. Currently, multimodal biomedical chatbots capable of accepting image inputs are scarce, with Visual Med-Alpaca\cite{2023vma} and LLaVA-Med\cite{li2023llava} being the only two known. Visual Med-Alpaca can accept image inputs and generate text dialogue outputs. Specifically, Visual Med-Alpaca processes input images through various image recognition models, combining the obtained visual information with textual prompts, enabling the model to generate appropriate responses. However, Visual Med-Alpaca's training on a limited biomedical dataset presents certain restrictions. Recently, Microsoft introduced LLaVA-Med, an end-to-end neural network model that processes images directly without the need for separate medical visual models. Furthermore, LLaVA-Med conducts experiments on a more diverse and larger dataset, demonstrating superior generalization and effectiveness.

\subsection{Biomedical Visual Question Answering}
The development of large models capable of autonomously generating answers to questions based on biomedical image inputs would greatly benefit physicians and patients alike, bearing significant implications for the enhancement of clinical diagnostic and treatment efficiency. Existing approaches to biomedical visual question answering principally fall into two categories: discriminative and generative methods.For discriminative methods, Visual Question Answering (VQA) is treated as a classification problem where the model seeks an answer from a predefined answer set\cite{eslami2023pubmedclip,he2020pathvqa,li2022self,zhang2023large}. Although this method can yield satisfactory performance, it operates within the constraints of a finite predefined answer dataset, thus requires adjustments or retraining when presented with a new or custom set of answers. Consequently, discriminative methods are not suited to the development of a biomedical chatbot capable of addressing open-ended queries.
To counter this limitation, researchers have pioneered generative methods that predict answers in the form of unrestricted text sequences\cite{bazi2023vision,liu2023q2atransformer,van2023open}. Generative methods not only accommodate custom answer sets, but also offer increased flexibility and yield more accurate responses.
A novel data generation method was proposed in \cite{li2023llava}, where GPT-4 was employed to generate multimodal instruction-following data pairs using medical image-text pairs from the publicly available dataset in PubMed Central \cite{zhang2023large}.

\subsection{Model Architecture}
There are mainly two methods for fine-tuning Large Language Models (LLMs) through biomedical visual question-answering dialogues. One approach involves individually fine-tuning the LLM on a medical dialogue dataset. The other method is the prefix tuning of language models (LMs), where a new trainable module connects the frozen image encoder and the causal LM, allowing for standard supervised fine-tuning of the entire model\cite{li2023llava,van2023open}. In \cite{van2023open}, a three-layer MLP network is employed to map visual features to the visual prefix. Furthermore, \cite{van2023open} incorporates a variety of pretrained LMs including GPT2-XL \cite{radford2019language}, BioMedLM \cite{venigalla2022biomedlm}, and BioGPT \cite{luo2022biogpt}, with sizes ranging from 1.5B to 2.7B. Conversely, \cite{li2023llava} employs linear projection alongside a 7B LM\cite{touvron2023llama}. In our model, ChatGLM \cite{zeng2022glm} was used as the LLM and was fine-tuned on an ophthalmology dialogue dataset. ChatGLM is an open-source dialogue language model that supports both English and Chinese conversations, boasting 6.2 billion parameters.

\section{Ophthalmology Dataset}
\label{sec:dataset}
Based on fundus medical imaging, the diagnosis of diseases can be obtained. In this study, we are currently able to provide diagnostic results for diabetic retinopathy, age-related macular degeneration, pathological myopia, glaucoma, and intraocular tumors, as well as lesion segmentation and staging for diabetic retinopathy. In order to maximize LLM's question-answering performance in specific diseases, we have devised two strategies to improve its performance.

\begin{figure*}
    \centering
	\includegraphics[scale=0.32]{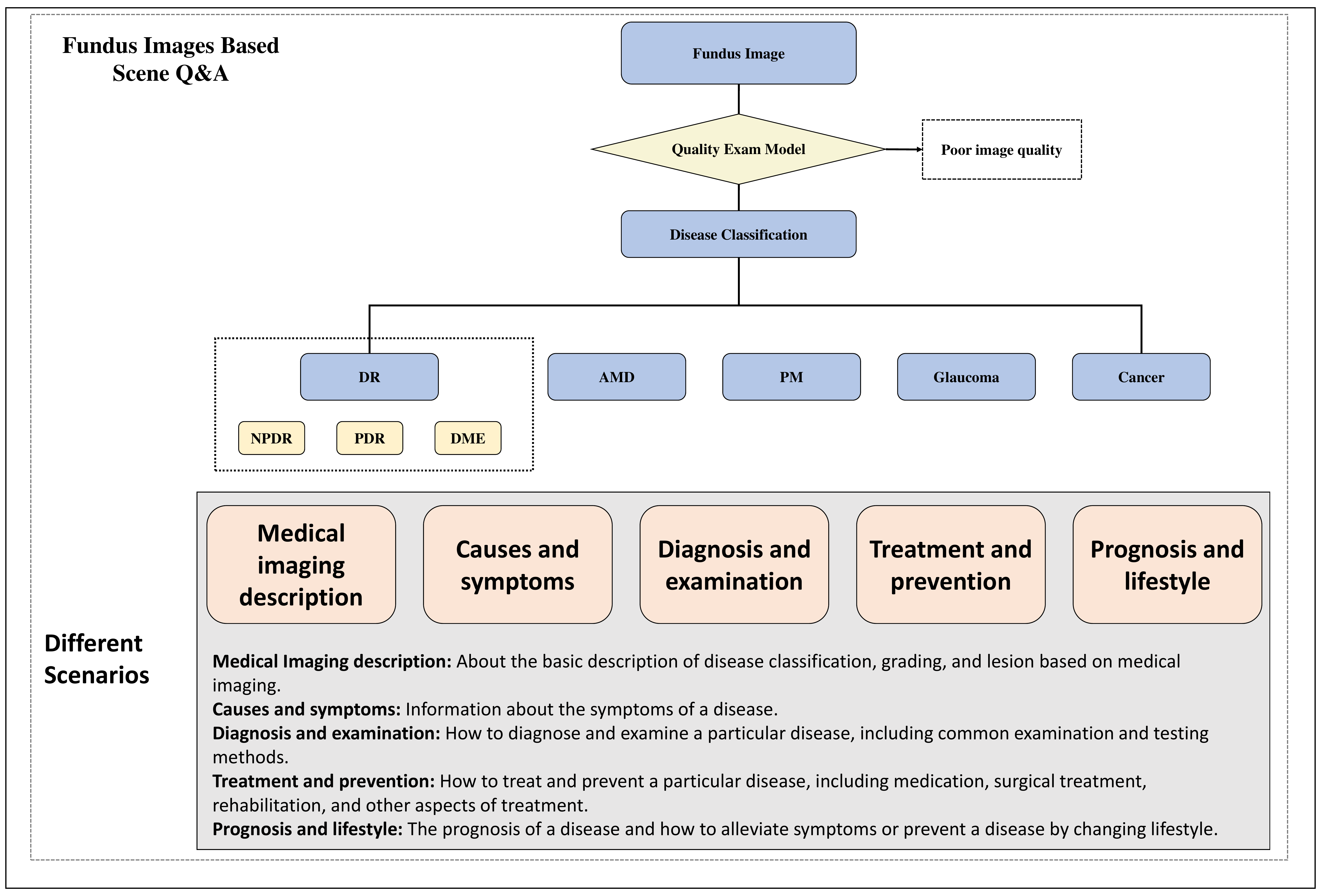}
	\caption{ {\bf Instruction fine-tuning datasets based on five different scenarios.} }
    \label{fig2}
\end{figure*}

\subsection{Instruction Data Construction}
Based on the open database crawler technology, we have constructed a large-scale medical knowledge graph with more than 100,000 nodes from Wikipedia, Dingxiangyuan and other sources. At the downstream of the Knowledge graph, we build the api parameters based on medical scenarios. Specifically, we construct instruction fine-tuning datasets based on five different scenarios, as is shown in Fig.\ref{fig2}. The knowledge based prompt example to create instructions is shown in appendix Fig.\ref{KGprompt}.

\textbf{Medical Imaging Description:} About the basic description of disease classification, grading, and lesion based on medical imaging.

\textbf{Causes and symptoms:} Information about the symptoms of a disease.

\textbf{Diagnosis and examination:} How to diagnose and examine a particular disease, including common examination and testing methods.

\textbf{Treatment and prevention:} How to treat and prevent a particular disease, including medication, surgical treatment, rehabilitation, and other aspects of treatment.

\textbf{Prognosis and lifestyle:} The prognosis of a disease and how to alleviate symptoms or prevent a disease by changing lifestyle.

\subsection{Medical Conversation Construction}
Instructions based on disease diagnosis and knowledge graph construction can be built around various scenarios of specific diseases using ChatGPT, leveraging additional knowledge to enable smaller LLMs to provide more realistic and robust medical Q\&A. However, the actual situation may be more complex.

The construction of knowledge instructions based on knowledge graphs cannot reflect the true concerns of patients. For example, patients generally have concerns about the heritability of the disease, medical costs, and efficacy. Instruction based fine-tuning can improve the authenticity of specific problems, but it cannot provide a better interactive experience.

In order to provide a dataset that is closer to real Q\&A scenarios and to create a better interactive experience, we proposed a dialogue prompt strategy based on real doctor-patient conversations. First, we extracted real conversations about ophthalmic diseases from the MedDialog dataset \cite{chen2020meddiag}, which contains genuine doctor-patient dialogues, as material. Then, we designed a set of prompts to enable ChatGPT to play the role of a medical expert, extracting patients' intentions from publicly available real doctor-patient dialogues and providing as professional and detailed medical explanations as possible. This approach not only enables the extraction of patients' intentions from real conversations, but also creates a more suitable dialogue dataset for patients.

\begin{figure*}
    \centering
    \resizebox{\textwidth}{!}{\includegraphics{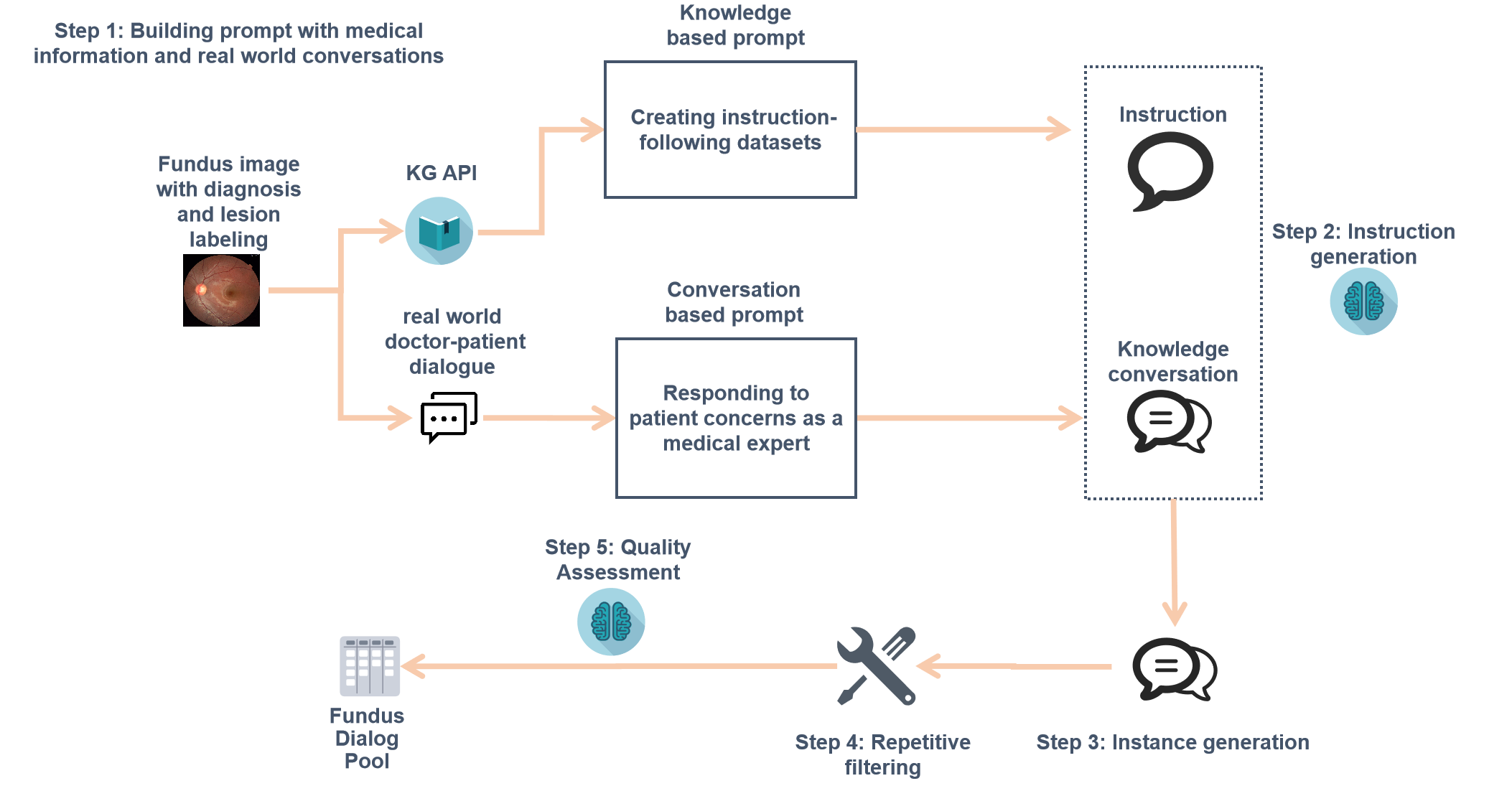}}
    \caption{ {\bf The process of building the fine-tuned Fundus dialog dataset.} }
    \label{fig1}
\end{figure*}

Our process for building the fine-tuned fundus dialog dataset is illustrated in Fig.\ref{fig1}. In step 1, prompts are created using both real-world medical-patient conversations and knowledge graphs. In step 2, medical knowledge based instructions and conversation are created using the ChatGPT interface. In step 3, data cleaning is performed to create instances. In step 4, duplicate data is removed by validating against existing datasets. In step 5, instance quality is evaluated using both manual review and GPT4. Finally, new instruction and conversation are added to fundus dialog pool.The construction of such a fine-tuned dataset not only enhances the authenticity and accuracy of LLM in medical knowledge but also creates a more friendly doctor-patient dialogue mechanism. The conversation based prompt example to create patient friendly conversation is shown in appendix Fig.\ref{Conprompt}.

Using these two strategies, we have constructed a Fundus Dialog Dataset consisting of over 20k instances related to ophthalmic diseases. Our experiments have shown that fine-tuning this dataset with ChatGLM can effectively improve the authenticity of disease-related Q\&A downstream from fundus medical imaging diagnosis. The fine-tuned OphGLM model possesses medical knowledge richness close to that of ChatGPT and can provide a more patient-friendly experience. Although the relevant data volume is not large, significant improvement has been achieved in LLM's factual answering ability.

\section{Methodology}
\label{sec:method}
\subsection{Overview}
\begin{figure*}
    \centering
	\includegraphics[scale=0.7]{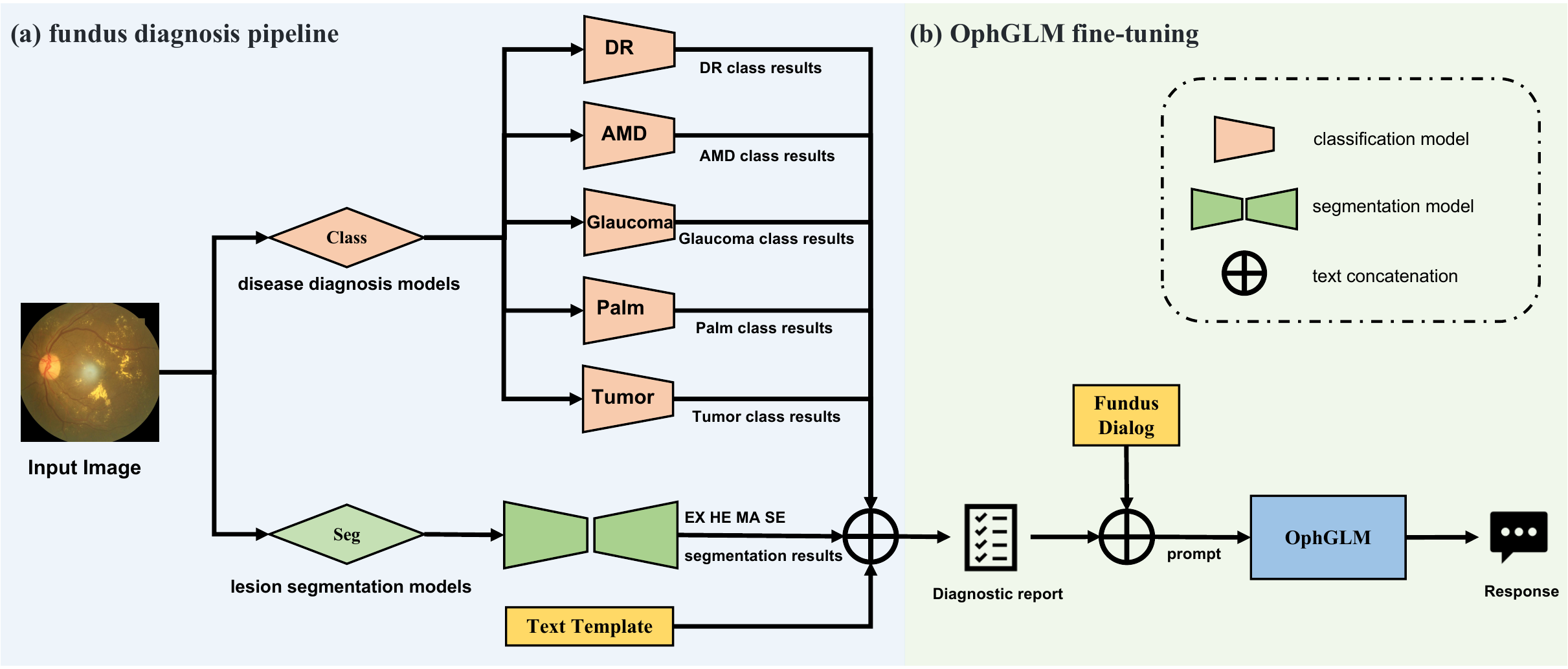}
	\caption{ {\bf Overall architecture of the proposed OphGLM.} (a) Fundus diagnosis pipeline. (b) OphGLM fine-tuning pipeline. }
    \label{fig3}
\end{figure*}

Our model, OphGLM, is a multimodal biomedical chatbot capable of accepting fundus images as input. OphGLM consists of two modules: the fundus diagnosis pipeline and the OphGLM pipeline. The overall pipeline of our proposed method is illustrated in Fig.\ref{fig3}.
Within the fundus diagnosis pipeline, there are two primary components: the fundus image classification segment and the fundus image segmentation segment. These segments work together to extract information from the input fundus images. The extracted visual information is integrated into structured text templates, forming diagnostic reports based on the input fundus images.
As for the OphGLM pipeline, it merges the fundus image diagnostic report with the fundus dialogue through a process of text concatenation. This forms a prompt that is then input into the OphGLM, ultimately generating a high-quality response.
\subsection{Fundus Diagnosis Pipeline}
In the fundus diagnosis pipeline of our model, the input fundus images are processed via a disease diagnosis module and a lesion segmentation module. Within the fundus image disease diagnosis module, we employ five different disease classification models, including a grading model for diabetic retinopathy (DR), an age-related macular degeneration (AMD) classification model, a glaucoma classification model, a pathological myopia classification model, and a fundus tumor classification model.
The specific functionalities of each disease classification model are as follows:
    The DR grading model can classify fundus images into normal, mild, non-proliferative, moderate non-proliferative, severe non-proliferative, and proliferative DR.
    The AMD classification model can classify fundus images into normal and AMD categories.
    The glaucoma classification model can classify fundus images into normal and glaucoma categories.
    The pathological myopia classification model can classify fundus images into normal and pathological myopia categories.
    The fundus tumor classification model can classify fundus images into normal and fundus tumor categories.
In the fundus image lesion segmentation module, we implemented segmentation of common fundus lesions, including hard exudates (EX), soft exudates (SE), microaneurysms (MA), and hemorrhages (HE).
Our visual models, based on Convolutional Neural Networks (CNN) and transformer architectures, simultaneously boast high accuracy and inference speeds. Once we have obtained the structured classification and segmentation results from these models, they are integrated into our predetermined text template, thus forming the diagnostic report for the fundus images.

\begin{table*}
	\footnotesize
	\centering	
	\caption{The datasets, number of images, and evaluation metrics used for the tasks in fundus image diagnosis model}
	\scalebox{1.05}{
			\begin{tabular}{c c c c c}
				\toprule
				 Task & Dataset & Number of images  & Evaluation Index \\
				\midrule
				\multirow{2}{*}{DR classification}  & DRD(Kaggle) & 88702 & Acc \\
				 &  DDR & 13673 & Acc  \\
                
				\multirow{2}{*}{glaucoma classification} & REFUGE  & 1200 & Acc \\
				 &ORIGA650  & 650  & Acc  \\
                
				PALM classification &BAIDU\_PALM  & 800   & Acc  \\
                
				AMD classification & IChallenge\_AMD400 & 400  & Acc  \\
                
				 tumor classification &Private Dataset & --- & Acc\\
				\midrule
				\multirow{3}{*}{lesion segmentation}    &   DDR & 757 &Dice  \\
				 & IDRID & 81 &Dice \\
				  & ADAM  & 400 & Dice  \\
				\bottomrule
	\end{tabular}}
	\label{table:1}
\end{table*}

\subsection{OphGLM Pineline}
During the training process of OphGLM, we fine-tuned the base model, ChatGLM\cite{zeng2022glm}, using a dataset consisting of over 20k fundus dialogues related to retinal diseases. As a result, we obtained OphGLM, which demonstrated a stronger capability in both disease diagnosis and question-answering.
Within the OphGLM pipeline, the fundus image diagnostic report and the user's input of fundus dialogue undergo text concatenation. By employing our strategically designed method, a dialogue prompt is generated and fed into OphGLM, which ultimately yields a high-quality response.

\section{Experiments}
\label{sec:experiment}
\subsection{Implement Details}
We provide a detailed overview of the training details for the fundus diagnosis pipeline. Our fundus diagnosis pipeline includes diabetic retinopathy classification, glaucoma classification, pathological myopia classification, age-related macular degeneration classification, fundus tumor detection, and diabetic retinopathy lesion segmentation.
The segmentation model for diabetic retinopathy lesions enables the segmentation of four types of lesions: hard exudates, soft exudates, microaneurysms, and hemorrhages. During the model training process, we utilized both publicly available datasets and private datasets to train and fine-tune the fundus diagnosis models. Specifically, the training dataset for the diabetic retinopathy lesion classification model is sourced from the Diabetic Retinopathy Detection dataset and DDR dataset\cite{li2019diagnostic}. The training dataset for glaucoma classification model is obtained from REFUGE and ORIGA650. The training dataset for pathological myopia classification model is from BAIDU\_PALM. The training dataset for age-related macular degeneration classification model is IChallenge\_AMD400. The training dataset for fundus tumor classification is from a private dataset of Beijing Tongren Hospital. The diabetic retinopathy lesion segmentation dataset is sourced from DDR, IDRID, and ADAM. The detailed information of the datasets is provided in Table \ref{table:1}. In this experiment, the evaluation metric for classification tasks is accuracy(ACC), and for segmentation tasks, it is dice coefficient(DSC).The equations of ACC and DSC are as follows:
\begin{equation}
ACC = \frac{TP + TN}{TP + TN + FP + FN}
\end{equation}

\begin{equation}
DSC = \frac{2 \times |X \cap Y|}{|X| + |Y|}
\end{equation}

For ACC equation, TP(True Positive) is the number of positive samples correctly classified as positive, TN(True Negative) is the number of negative samples correctly classified as negative, FP(False Positive) is the number of negative samples incorrectly classified as positive, and FN(False Negative) is the number of positive samples incorrectly classified as negative.
For DSC equation, $|X \cap Y|$ represents the size of the intersection between the prediction results (typically the segmented image) and the actual results (typically the manually annotated image). $|X|$ and $|Y|$ represent the sizes of the prediction and actual results, respectively.

\subsection{Peformance}
The performance of each model in our fundus diagnosis pipeline is shown in Table \ref{table:2}. The visual input example of a DR fundus image is shown in appendix Fig.\ref{visualinput} and some medical knowledge based Q\&A examples are shown in appendix Table \ref{table:3} and Table \ref{table:4}.
% \begin{table}[htb]
% \centering
% \caption{Model performance}
% \begin{tabular}{llll}
% \toprule
% Algorithm Type & Algorithm & Acc & Dice \\
% \midrule
% classification & DR\_class\_model & 0.970 & / \\
%  & Glaucoma\_class\_model & 0.940 & / \\
%  & PALM\_calss\_model & 0.998 & / \\
%  & AMD\_class\_model & 0.984 & / \\
%  & Tumor\_class\_model & 0.999 & / \\
% \hline
% segmentation & EX & / & 0.854 \\
%  & HE & / & 0.805 \\
%  & MA & / & 0.699 \\
%  & SE & / & 0.812 \\
% \bottomrule
% \end{tabular}
% \label{table:2}
% \end{table}
\begin{table}
\centering
\caption{The performance of models in fundus diagnosis pipeline.}
\begin{tabular}{c c c c}
\toprule
Algorithm Type & Algorithm & Acc & Dice \\
\midrule
\multirow{5}{*}{classification} & DR\_class\_model & 0.970 & -- \\
 & Glaucoma\_class\_model & 0.940 & -- \\
 & PALM\_calss\_model & 0.998 & -- \\
 & AMD\_class\_model & 0.984 & -- \\
 & Tumor\_class\_model & 0.999 & -- \\
\hline
\multirow{4}{*}{segmentation} & EX\_seg\_model & -- & 0.854 \\
 & HE\_seg\_model & -- & 0.805 \\
 & MA\_seg\_model & -- & 0.699 \\
 & SE\_seg\_model & -- & 0.812 \\
\bottomrule
\end{tabular}
\label{table:2}
\end{table}

\section{Conclusions}
\label{sec:conclusion}
We have developed OphGLM, a large-scale language and vision assistant for ophthalmology based on instructions and dialogue. Our model was fine-tuned on a dataset of over 20k instructional fine-tuning and dialogue dataset, carefully crafted based on real-world conversations and medical knowledge. This process provided OphGLM with powerful question answering abilities tailored specifically for ophthalmology.

Despite significant progress in ophthalmic visual dialogue, there is still room for improvement in our model. We will continue to conduct research and generate more high-quality image-language data to enhance the overall capabilities of the model. In addition, we plan to enrich and strengthen the upstream diagnostic models, providing multi-label disease classification and lesion annotation capabilities. Finally, we plan to incorporate other medical imaging modalities in ophthalmology, such as OCT, into the model to create a comprehensive medical-assistant system.

%%%%%%%%% REFERENCES
{\small
\bibliographystyle{ieee_fullname}
\bibliography{ref}
}

\section{Appendix}

\begin{figure*}
\centering
\resizebox{\textwidth}{!}{\includegraphics{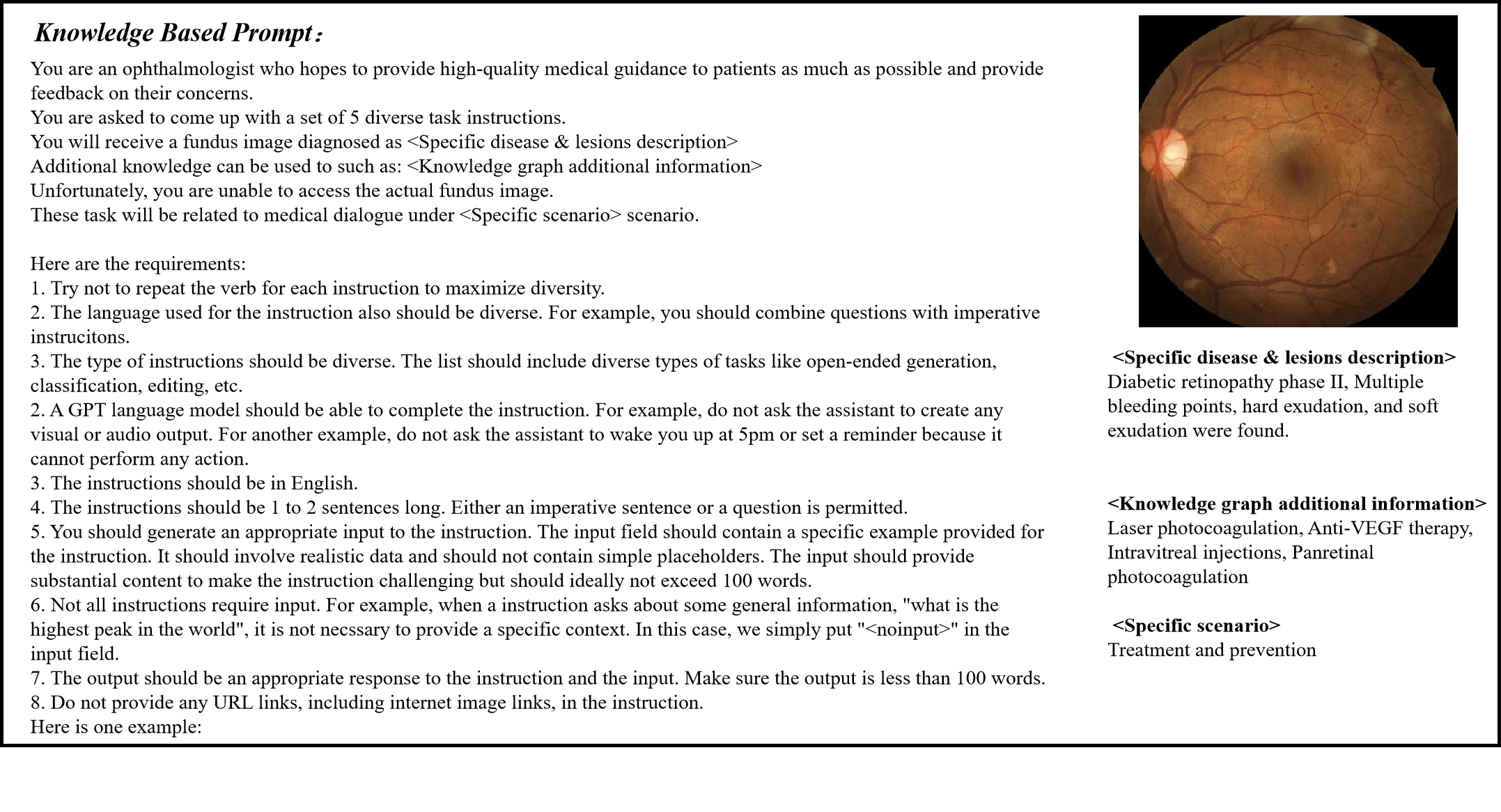}}
\caption{ {\bf Knowledge Based Prompt.} }
\label{KGprompt}
\end{figure*}

\begin{figure*}
\centering
\resizebox{\textwidth}{!}{\includegraphics{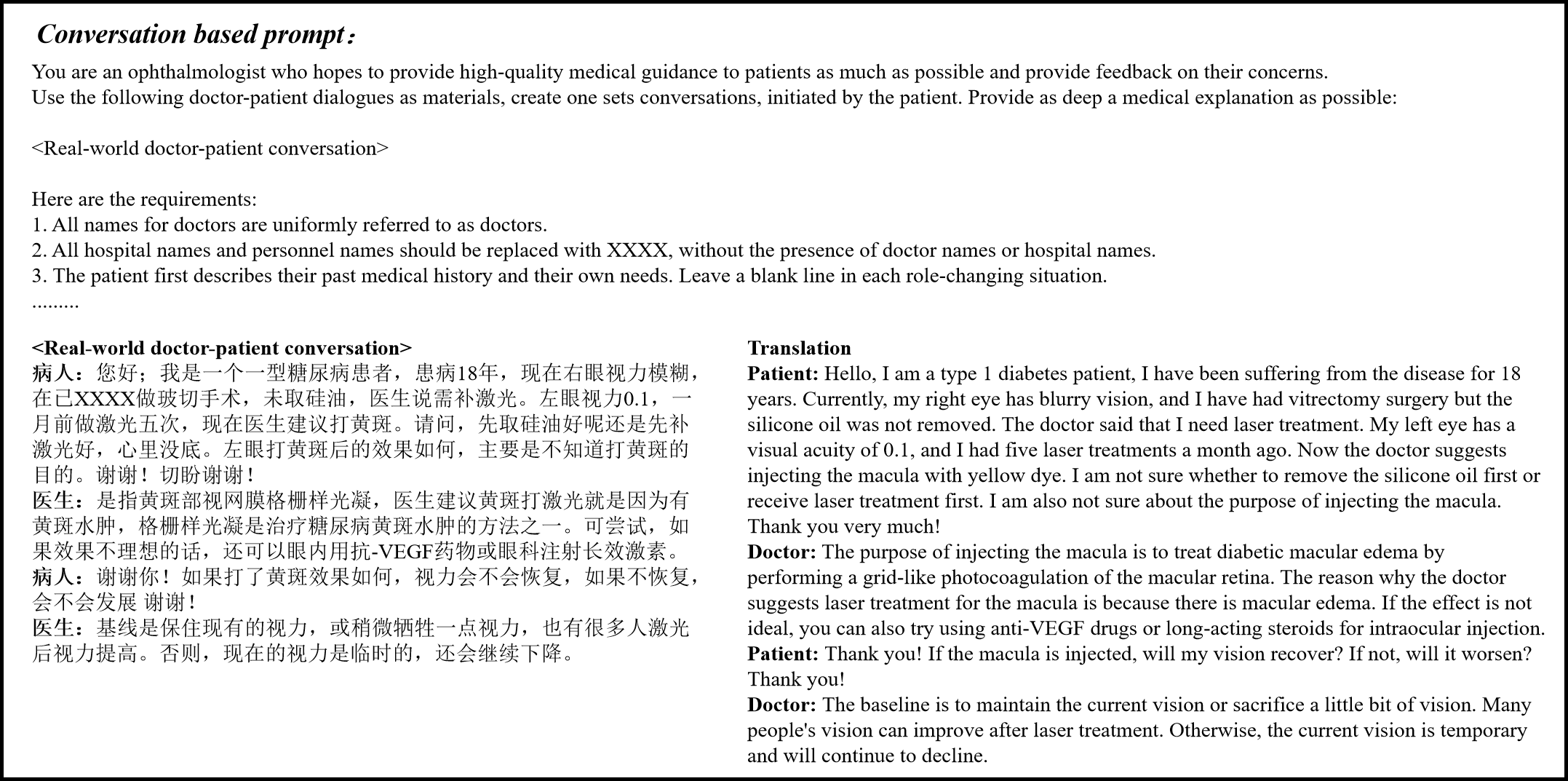}}
\caption{ {\bf Conversation Based Prompt.} }
\label{Conprompt}
\end{figure*}

\begin{figure*}
\centering
\resizebox{\textwidth}{!}{\includegraphics{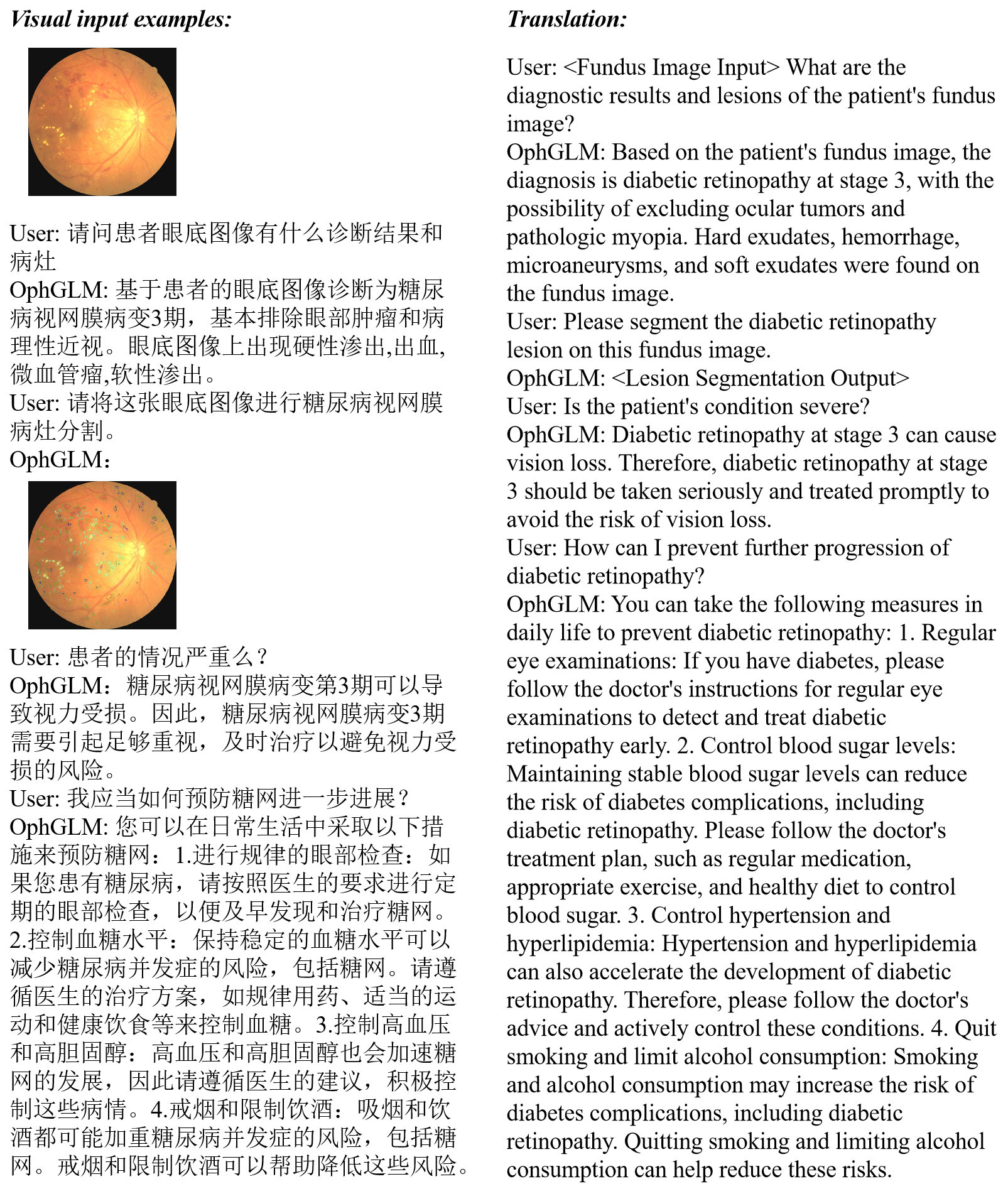}}
\caption{ {\bf Visual Input Example.} }
\label{visualinput}
\end{figure*}

\begin{table*}
	\footnotesize
	\centering	
	\caption{Improving the ability to answer medical basic questions}
	\scalebox{1.01}{
			\begin{tabular}{c |p{0.21\textwidth} p{0.21\textwidth} p{0.21\textwidth}}
				\hline
				 & example 1 & example 2 & example 3 \\
				\hline
				% Fundus Image  & \includegraphics[width=0.26\textwidth]{fig/amd.png} 
    %             & \includegraphics[width=0.26\textwidth]{fig/amd2.png}   \\
                
			   Ralated Disease  & AMD & AMD & DR \\
                  \hline
				 \multirow{4}{*}{Question} & What kind of disease is AMD?
                 & Is age-related macular degeneration a common disease? What is the incidence rate among people over 50 years old?   
                 & What are the lesions in the fundus of diabetic retinopathy?\\
                
				% ChatGPT &AMD (Age-related Macular Degeneration) is a common degenerative disease that occurs in the eyes of older people. It causes the central area of the retina to lose sensitivity, thus affecting central vision. This disease can cause visual impairment and blindness, bringing serious physical and psychological burden to patients. Therefore, AMD can be classified into the category of diseases as a common retinal disease.  
    %             & Age-related Macular Degeneration (AMD) is a common retinal disease that occurs in the elderly, mainly in people over 50 years old. According to statistical data, the proportion of people over 50 years old with AMD can be as high as 10\% to 25\%. It should be noted that the incidence of AMD gradually increases with age. Specifically, in the population over 60 years old, the incidence of AMD is higher, reaching 30\% to 40\%. In addition, AMD is also related to other factors such as genetics, smoking, high blood pressure, and high cholesterol. If you suspect that you have AMD, it is recommended to seek medical attention promptly and receive professional diagnosis and treatment.  \\
                \hline
				\multirow{7}{*}{ChatGLM} & AMD is a variant of AMD, also known as Houssay-Holopigian syndrome.
                & The prevalence of age-related macular degeneration in people over the age of 50 is approximately 10\% according to data from the American Heart Association. 
                & The following retinal lesions may appear in the fundus images of diabetic retinopathy: retinal neovascularization, retinal detachment, retinal nerve fiber degeneration, and retinal nerve fiber inflammation.\\
                
				\hline
				\multirow{39}{*}{OphGLM}    &   AMD(Age-related Macular Degeneration) is a type of retinal disease which mainly manifests as lesions in the macular region of the retina. It is a chronic and progressive disease that usually occurs with increasing age, and is more common in people over 60 years old. 
               & Age-related Macular Degeneration (AMD) is a common macular disease that typically affects people over the age of 50. According to the World Health Organization, the global prevalence of AMD significantly increases with age, particularly in Asia and South America. According to the International Age-Related Macular Degeneration Foundation's data in 2019, the global prevalence of AMD is around 20.6\%, with a rate of about 30\% for those aged over 50. However, the prevalence may vary in different countries and regions depending on factors such as population structure, lifestyle habits, and medical conditions. AMD is a common macular disease that can lead to vision loss and visual impairment. To prevent AMD, people over the age of 50 are advised to take active measures such as regular physical examinations, maintaining a healthy lifestyle, and protecting their eyes. 
               & Diabetic retinopathy is a disease caused by damage to the retinal blood vessels due to diabetes, and is common in people with diabetes. Diabetic retinopathy mainly includes the following lesions:
                \begin{enumerate}
      
               \item Microvascular abnormalities, including microaneurysms, narrowing, and abnormal tortuosity. 
               \item Hard exudates: due to damage to the retinal capillaries and neurons, fluid, protein, and other substances in the blood will penetrate into the retinal area, forming hard exudates. 
               \item Non-proliferative diabetic retinopathy: including microvascular embolism, punctate hemorrhage, exudation and crystalline exudation. 
               \item Proliferative diabetic retinopathy: including neovascularization and scar tissue, these lesions may extend into the vitreous cavity and may cause retinal detachment. 
               \end{enumerate}
               Diabetic retinopathy has various types of lesions and degrees of severity.If you have diabetes, please pay close attention to eye health, undergo regular eye exams, and follow your doctor's advice for treatment and preventive measures, such as controlling blood sugar, blood pressure, and lipids. Early detection and treatment of lesions can prevent or slow down their progression and reduce the risk of eye complications.\\
				\hline
	\end{tabular}}
	\label{table:3}
\end{table*}

\begin{table*}
	\footnotesize
	\centering	
	\caption{Improving the ability to answer medical basic questions}
	\scalebox{1.01}{
			\begin{tabular}{c |p{0.21\textwidth} p{0.21\textwidth} p{0.21\textwidth}}
				\hline
				 & example 4 & example 5 & example 6 \\
				\hline
				% Fundus Image  & \includegraphics[width=0.26\textwidth]{fig/amd.png} 
    %             & \includegraphics[width=0.26\textwidth]{fig/amd2.png}   \\
                
			   Ralated Disease  &DR & Glaucoma & PM \\
                  \hline
				 \multirow{3}{*}{Question} & How can I prevent further progression of diabetic retinopathy if I already have diabetes?
                 & What medications can be used to treat or slow the progression of glaucoma?   
                 & What are the lesions of pathologic myopia in fundus images?\\
                \hline
				\multirow{8}{*}{ChatGLM} & To prevent diabetic retinopathy, it is necessary to control diet and exercise, and regularly monitor blood sugar levels.
                & Medications used to treat glaucoma include Kaiyipin and other drugs, which can reduce the accumulation of aqueous humor and lower the risk of glaucoma by lowering intraocular pressure (IOP) and promoting the outflow of aqueous humor. 
                & The fundus images of pathological myopia may show retinal detachment, choroidal detachment, and retinal neovascularization.\\
                
				\hline
				\multirow{43}{*}{OphGLM}    &   Diabetes is one of the main risk factors for developing diabetic retinopathy. If you have been diagnosed with diabetes, you can take the following measures in your daily life to prevent diabetic retinopathy: 
                \begin{enumerate}

                \item Undergo regular eye exams: if you have diabetes, please undergo regular eye exams as required by your doctor to detect and treat diabetic retinopathy early. 
                \item Control blood sugar levels: maintaining stable blood sugar levels can reduce the risk of diabetes complications, including diabetic retinopathy. Please follow your doctor's treatment plan, such as taking medication regularly, appropriate exercise, and a healthy diet to control blood sugar. 
                \item Control high blood pressure and high cholesterol: high blood pressure and high cholesterol can also accelerate the development of diabetic retinopathy, so please follow your doctor's advice and actively control these conditions. 
                \item Quit smoking and limit alcohol consumption: smoking and alcohol consumption can both increase the risk of diabetes complications, including diabetic retinopathy. Quitting smoking and limiting alcohol consumption can help reduce these risks. 
                \end{enumerate}
               & Glaucoma is a disease characterized by elevated intraocular pressure, which cannot be cured on its own and typically requires medication to control eye pressure and prevent further vision damage. Here is a list of some drugs used to treat glaucoma: 
               \begin{enumerate}

               \item Intraocular pressure inhibitors such as Reye (latanoprost) can reduce intraocular pressure by inhibiting bacterial growth and are suitable for glaucoma patients. 
               \item Anti-glaucoma drugs such as AstraZeneca (brinzolamide) and Novartis (brimonidine) and other brands can slow down the progression of glaucoma and loss of vision by reducing intraocular pressure. 
               \item Over-the-counter eye pressure-lowering drugs such as 0.1\% atropine eye drops can relax the retinal nerve fibers and vascular bundles to lower intraocular pressure and are suitable for glaucoma patients. 
               \item Combination therapy of intraocular pressure inhibitors and anti-glaucoma drugs can increase treatment efficacy and reduce the risk of complications. These medications should be used under the guidance of a doctor, as different medications may have different side effects and adverse reactions. 
               \end{enumerate}
               Additionally, glaucoma medication is just one method of controlling intraocular pressure and patients need to take other measures, such as avoiding eye strain, undergoing regular vision checks, and following their doctor's recommendations.
               & Pathological myopia, also known as high myopia or degenerative myopia, is a common type of myopia characterized by an excessively elongated eyeball that causes light to not focus fully on the retina, leading to retinal damage. Here are some common retinal lesions that can be seen in pathological myopia: 
               \begin{enumerate}

               \item Retinal detachment: In pathological myopia, the retina may become detached due to excessive stretching and pressure, which can lead to vision loss. 
               \item Neovascularization: Due to the elongated eyeball in pathological myopia, new blood vessels may grow in the retina, causing malnutrition and bleeding. 
               \item Fissures: The elongated eyeball in pathological myopia can cause fissures in the retina, which can lead to retinal rupture and vision loss. 
               \item Choroidal neovascularization: Excessive stretching and pressure on the retina in pathological myopia can lead to choroidal neovascularization, a common retinal disease that can result in vision loss. 
               \item Leopard-spot-like vision: Due to retinal damage caused by excessive stretching and pressure in pathological myopia, leopard-spot-like vision patterns may appear in the retina. 
               \end{enumerate}
               These are common retinal lesions associated with pathological myopia, and if you have any questions, it is recommended to consult a doctor for more detailed advice and diagnosis.\\
				\hline
	\end{tabular}}
	\label{table:4}
\end{table*}

\end{document}